%% file: main.tex
\documentclass[letterpaper,10pt,conference]{ieeeconf}  

\IEEEoverridecommandlockouts                              
\usepackage{graphicx} 
\usepackage{amsmath} 
\usepackage{amssymb}  
\usepackage{algorithm}
\usepackage[colorlinks=true,linkcolor=blue,citecolor=blue]{hyperref}
\usepackage{algorithmic}
\usepackage{xcolor}
\usepackage{soul}
\usepackage{multirow}
\usepackage{makecell}
\usepackage{hyperref}
\hypersetup{%
  colorlinks=true,
  linkcolor=blue,
  linkbordercolor={0 0 1}
}
\makeatletter
\let\NAT@parse\undefined
\makeatother
\usepackage{subcaption}
\usepackage{float}
\graphicspath{{figures/}}

\title{\LARGE \bf Cooking Object's State Identification Without Using Pretrained Model}

\author{Md Sadman Sakib\\
Department of Computer Science and Engineering \\ University of
South Florida \\
mdsadman@usf.edu
}

\begin{document}

\maketitle

\thispagestyle{empty}
\pagestyle{empty}


\begin{abstract}

Recently, Robotic Cooking has been a very promising field. To execute a recipe, a robot has to recognize different objects and their states. Contrary to object recognition, state identification has not been explored that much. But it is very important because different recipe might require different state of an object. Moreover, robotic grasping  depends on the state. Pretrained model usually perform very well in this type of tests. Our challenge was to handle this problem without using any pretrained model. In this paper, we have proposed a CNN and trained it from scratch. The model is trained and tested on the dataset from cooking state recognition challenge. We have also evaluated the performance of our network from various perspective. Our model achieves 65.8\% accuracy on the unseen test dataset. 

\end{abstract}



\input{intro}

\input{preprocessing}

\input{methods}

\input{evaluation}

\input{discussion}

\bibliographystyle{unsrt}
\bibliography{ref}

\end{document}

%% file: intro.tex
\section{Introduction}
The goal of Artificial Intelligence is to build robots that can perform sophisticated tasks for human. Robotic cooking is one such field where AI meets the physical world. A Robotic Cook can be a great solution to the elderly people or people with disabilities who have problems to prepare their food. To execute a recipe, it is not enough to recognize the objects. It is also important to recognize the object's state to understand which form of ingredients to use, when the cooking is done etc. Additionally, robotic grasping style and motion of the end effectors might also vary depending on the object's state. For example, picking a whole onion require different grasping from picking diced onion. Though object recognition is very well explored, state classification has not got that much attention.

Functional Object Oriented Network (FOON) \cite{Paulius2016FunctionalON, Paulius2018FunctionalON, Paulius2019ASO, Paulius2020TaskPW}, a knowledge network have integrated state with object and manipulation motion. It produces a sequence of tasks that can lead a robot to successful execution of a recipe. From this graphical model, robot can learn about different states of an object. State recognition also plays a vital role in choosing the appropriate grasping motion. Shape matching algorithm \cite{grasp} is an attempt to solve this problem. 

Object–object-interaction affordance knowledge \cite{Sun2014ObjectobjectIA} is also another approach in this line of research. The interactive motions between paired objects in a human–object–object way instead of considering a single object. The learned knowledge is represented as Bayesian Network. It improves the recognition reliability.

Most recently, there have been some other approaches particularly designed to solve state classification challenge \cite{Sharma2018StateCW,Paul2018ClassifyingCO, Jelodar2018Identifying, Sun2018AIMP}. These models are based on pretrained models like VGG\cite{vgg}, ResNet\cite{Resnet}, Inception network \cite{googlenet}. The current state recognition challenge requires to classify the states without using any pretrained model. That is why we have fine-tuned our parameters by thorough analysis and tried to find the best configuration for our case. Then, we have trained the model from scratch. 

The dataset contains 7210 training and 1543 validation images with 11 states. The images are related to cooking ingredients. Since the dataset is small, we have used several augmentation techniques. The model contains six convolutional layers, two fully connected layers, batch normalization layers and max pooling layers. In the next section, we will discuss about the data collection and preprocessing steps. In the following sections, we will present our model and evaluate its performance from various perspective.  

%% file: preprocessing.tex
\section{Data Collection and Preprocessing}

\begin{figure}[h]
	\centering
	\includegraphics[width=\linewidth, height=6cm]{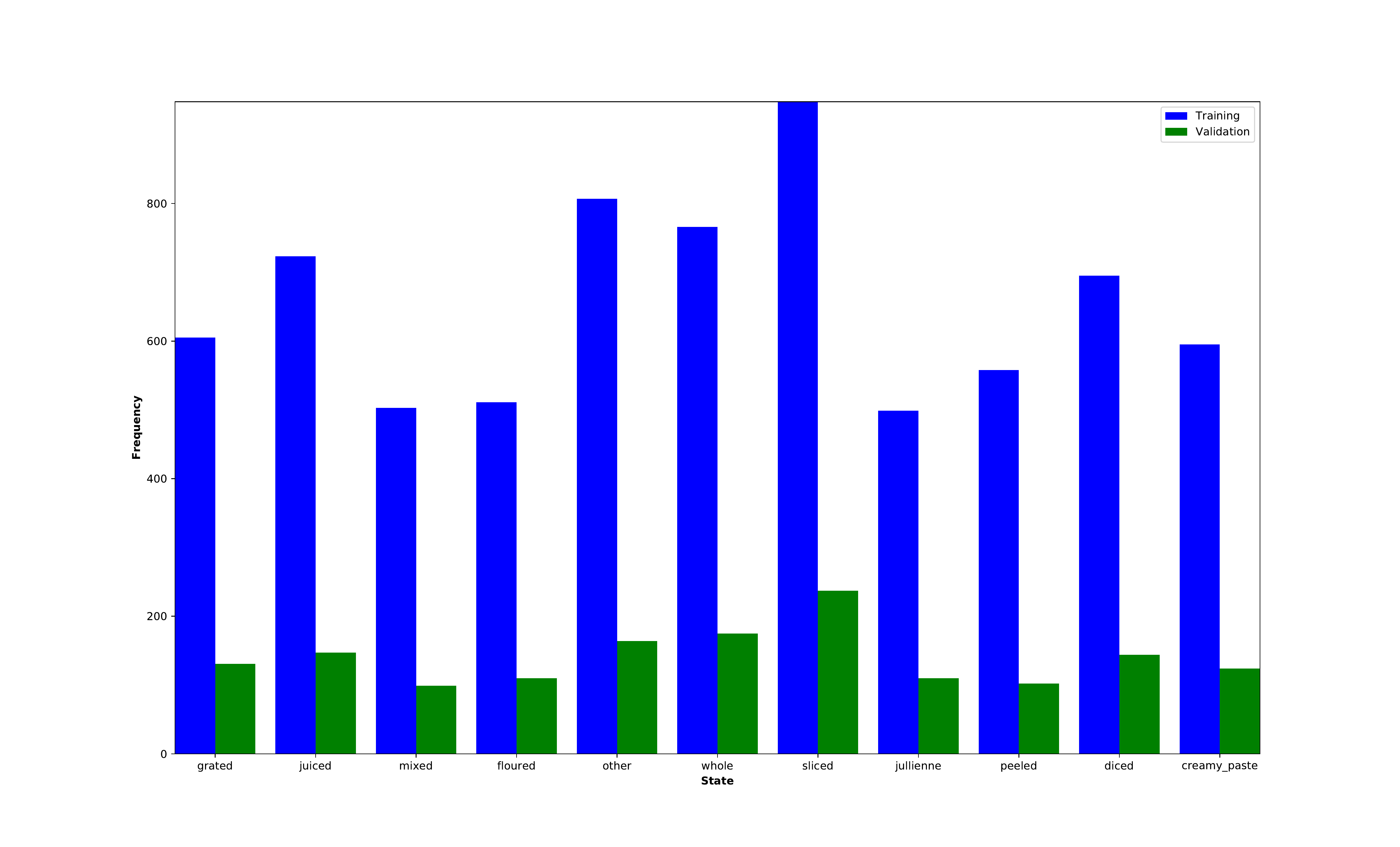}
 	\caption{Frequency of states in training and validation dataset}
	\label{fig:freq}
\end{figure}

We have used the dataset version 2.0 used in cooking state recognition challenge \cite{Jelodar2018Identifying, Jelodar2019JointOA}. Additionally, students were asked to annotate more images with appropriate state. The final dataset for this project was preapared by combining these two set of labelled images. The training and validation set contain 7210 and 1543 images respectively. Rest of the images were kept hidden for testing. The dataset has 11 cooking states. Figure \ref{fig:freq} shows the frequency of states in training and validation dataset.

\begin{figure}[h]
  \begin{subfigure}[b]{0.12\textwidth}
    \includegraphics[width=\textwidth]{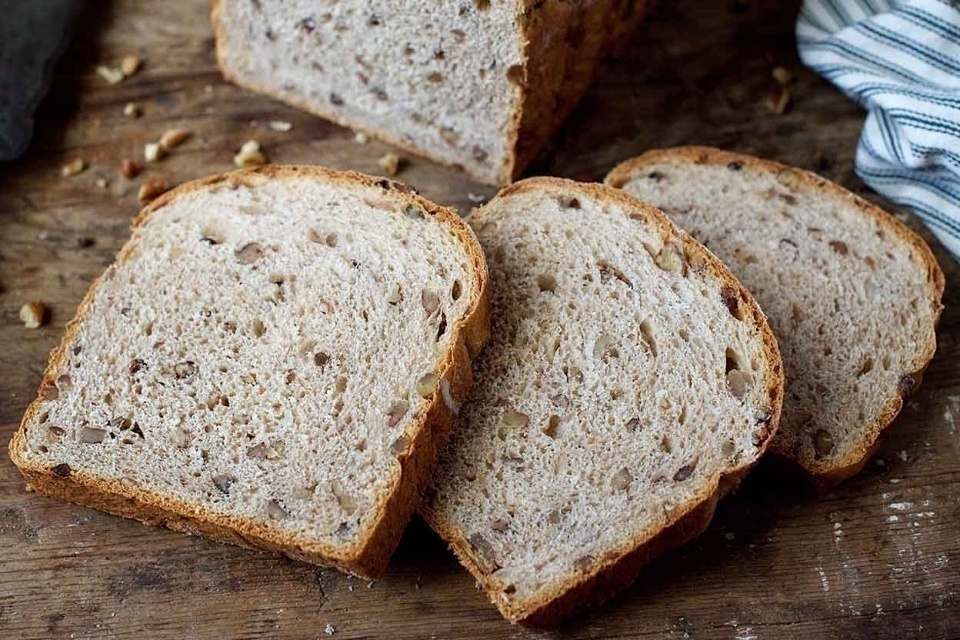}
    \caption{}
    \label{fig:1}
  \end{subfigure}\hfill
  \begin{subfigure}[b]{0.12\textwidth}
    \includegraphics[width=\textwidth]{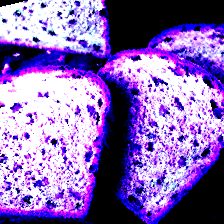}
    \caption{}
    \label{fig:2}
  \end{subfigure}\hfill
  \begin{subfigure}[b]{0.12\textwidth}
    \includegraphics[width=\textwidth]{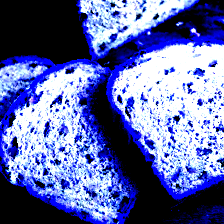}
    \caption{}
    \label{fig:3}
  \end{subfigure}\hfill
  \begin{subfigure}[b]{0.12\textwidth}
    \includegraphics[width=\textwidth]{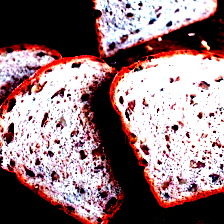}
    \caption{}
    \label{fig:4}
  \end{subfigure}\hfill
  \begin{subfigure}[b]{0.12\textwidth}
    \includegraphics[width=\textwidth]{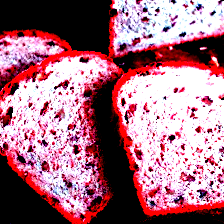}
    \caption{}
    \label{fig:5}
  \end{subfigure}\hfill
  \begin{subfigure}[b]{0.12\textwidth}
    \includegraphics[width=\textwidth]{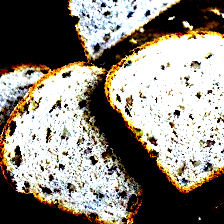}
    \caption{}
    \label{fig:6}
  \end{subfigure}\hfill
  \begin{subfigure}[b]{0.12\textwidth}
    \includegraphics[width=\textwidth]{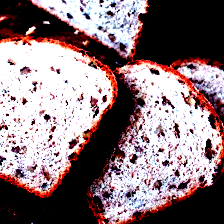}
    \caption{}
    \label{fig:7}
  \end{subfigure}\hfill
  \begin{subfigure}[b]{0.12\textwidth}
    \includegraphics[width=\textwidth]{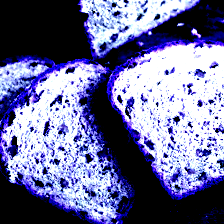}
    \caption{}
    \label{fig:8}
  \end{subfigure}\hfill
  \begin{subfigure}[b]{0.12\textwidth}
    \includegraphics[width=\textwidth]{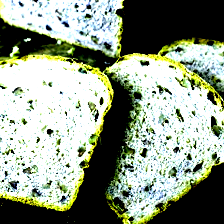}
    \caption{}
    \label{fig:1}
  \end{subfigure}\hfill
  \begin{subfigure}[b]{0.12\textwidth}
    \includegraphics[width=\textwidth]{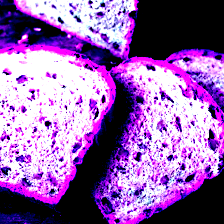}
    \caption{}
    \label{fig:9}
  \end{subfigure}\hfill
  \begin{subfigure}[b]{0.12\textwidth}
    \includegraphics[width=\textwidth]{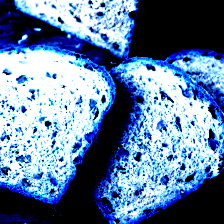}
    \caption{}
    \label{fig:10}
  \end{subfigure}\hfill
  \begin{subfigure}[b]{0.12\textwidth}
    \includegraphics[width=\textwidth]{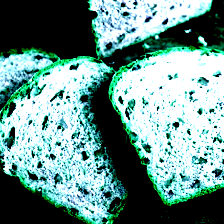}
    \caption{}
    \label{fig:11}
  \end{subfigure}\hfill
  \caption{(a) Original image, (b-l) Augmented images}
  \label{fig:augmentation}
\end{figure}

Though, figure \ref{fig:freq} indicates that the dataset is well balanced, it is a very small dataset. We have handled this problem by using augmentation techniques such as: rotation, shifting, cropping, flipping etc. As a result, in each epoch, we get a new transformed image. \ref{fig:augmentation} presents the original image and its augmented images collected from the first 11 epochs. We normalize the images before feeding it to the neural network for faster convergence.

%% file: methods.tex
\section{Methodology}

\footnote{The full-fledged implementation is available at: \url{https://github.com/sadman3/state-classification}}.

A complex deep learning network usually suffers from overfitting if the training set is small. In that case, it tries to create a one-to-one mapping with the training data which leads to high variance. Pretrained models, having a fairly complex structure, perform well on small dataset because they are already trained on a huge dataset. Since we were not allowed to use any pretrained model, we have designed a very simple network to train from scratch. In the following subsections we will discuss our proposed model and selected values of hyperparameters. 

\subsection{Proposed Model}
\begin{figure*}[ht]
	\centering
	\includegraphics[width=\linewidth]{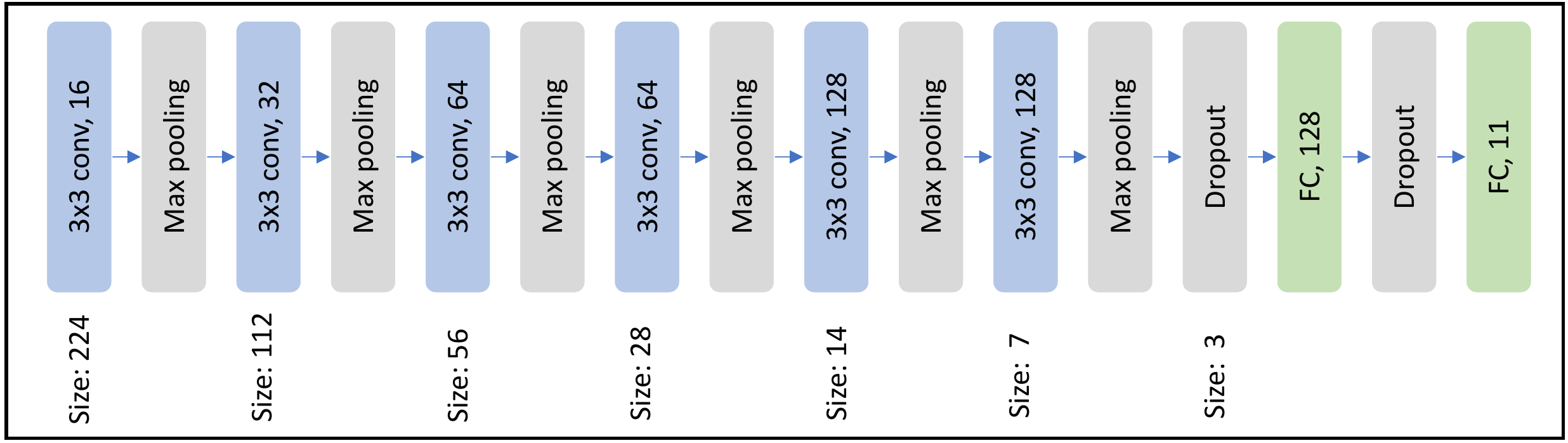}
 	\caption{Proposed Convolutional Neural Network to identify object's state}
	\label{fig:model}
\end{figure*}

\begin{figure}[ht]
	\centering
	\includegraphics[width=\linewidth]{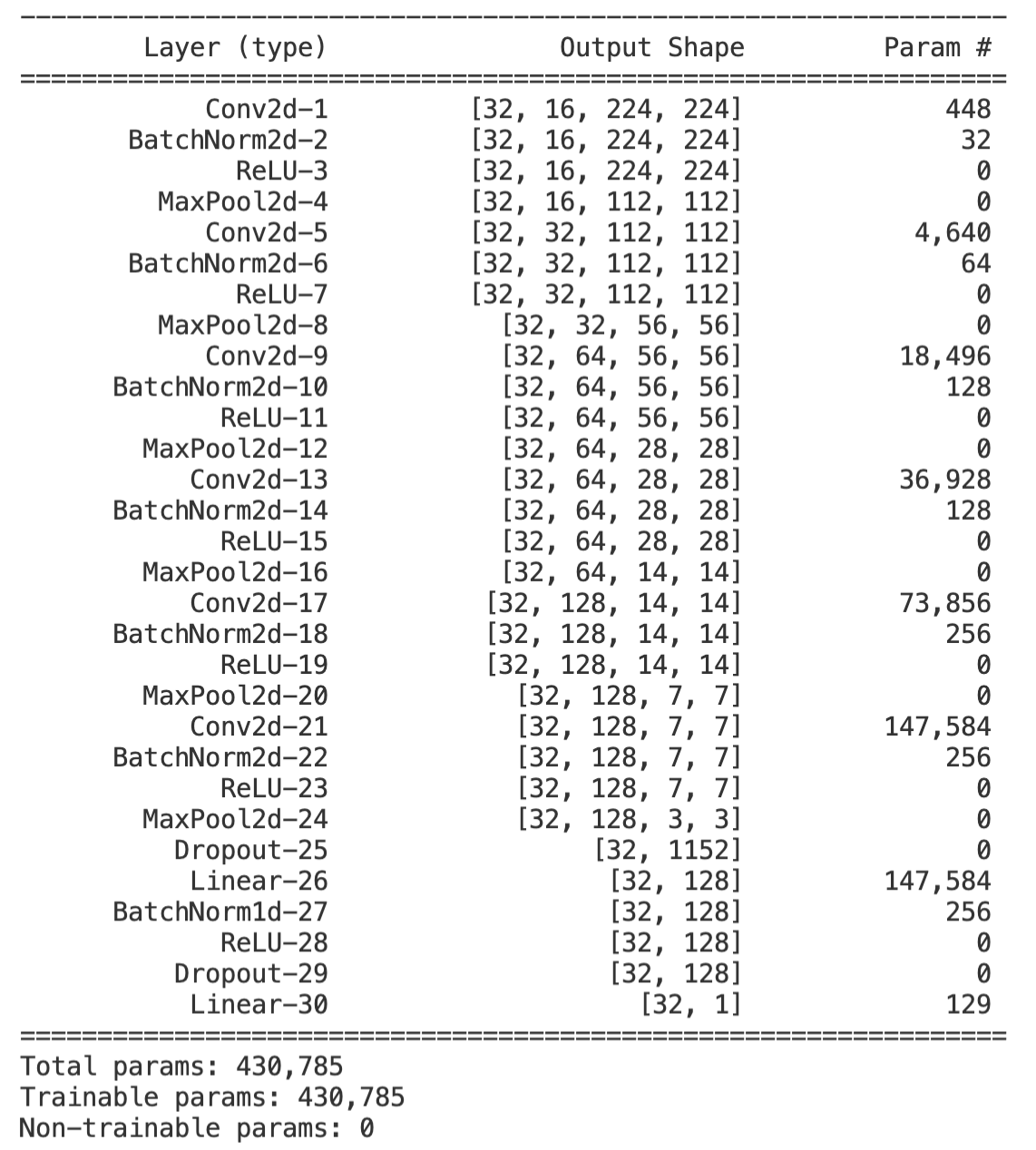}
 	\caption{Model summary}
	\label{fig:params}
\end{figure}

Figure \ref{fig:model} presents our model. It contains six convolutional layers and two fully connected layers. We have used ReLU as the activation function.  All the convolutional filters are of $3*3$ size. After each convolutional layer, there is a max pooling layer with a filter of size 2 and stride 2. It reduces the size of the image by 2. Our goal was to reduce the size of the input before feeding it to the fully connected layer so that the number of parameter keeps low. We have total $430,785$ trainable parameters in our network. We also have a batch normalization layer after each convolutional layer. Since, the number of state is 11, the softmax layer has 11 units as output. Figure \ref{fig:params} shows the output shape and number of trainable parameters in each layer of our model.

\subsection{Hyperparameters}
Table \ref{table:hyperparams} presents the hyperparameter values selected for our model. 
\begin{table}[ht]
\centering
\captionof{table}{Hyperparameter values selected for our model}
\begin{tabular}{|c|c|}
\hline
\textbf{Hyperparameter} & \textbf{Selected value} \\ \hline
Optimizer & SGD  \\ \hline
Momentum & $0.9$  \\ \hline
Dropout factor  & $0.5$ \\ \hline
Batch size & $32$ \\ \hline
Number of epochs & $80$ \\ \hline
Learning rate & $0.01$ \\ \hline
Learning rate decay & \makecell{first $50$ epochs: fixed \\ After that: $10\%$ decay after every 10 epoch } \\ \hline
\end{tabular}
\label{table:hyperparams}
\end{table}

%% file: evaluation.tex
\section{Evaluation and Results}

In this section, we will discuss how we have chosen our hyperparamters, some optimization techniques and the overall performance of our model. 

\subsection{Optimizer}

\begin{figure}[ht]
  \begin{subfigure}[b]{0.24\textwidth}
    \includegraphics[width=\textwidth]{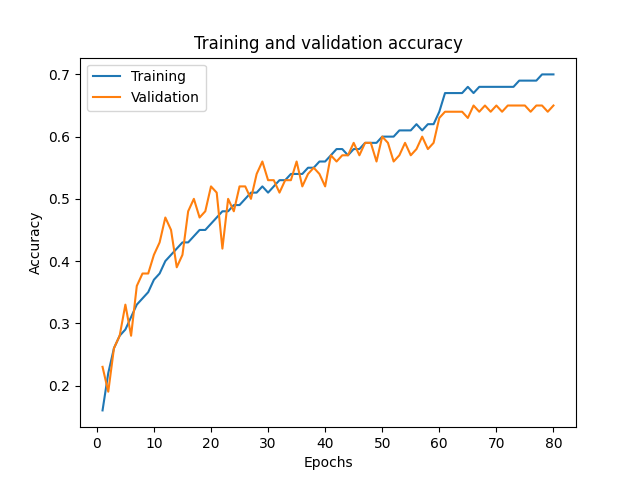}
    \caption{SGD, batch size 32}
    \label{fig:01}
  \end{subfigure}\hfill
  \begin{subfigure}[b]{0.24\textwidth}
    \includegraphics[width=\textwidth]{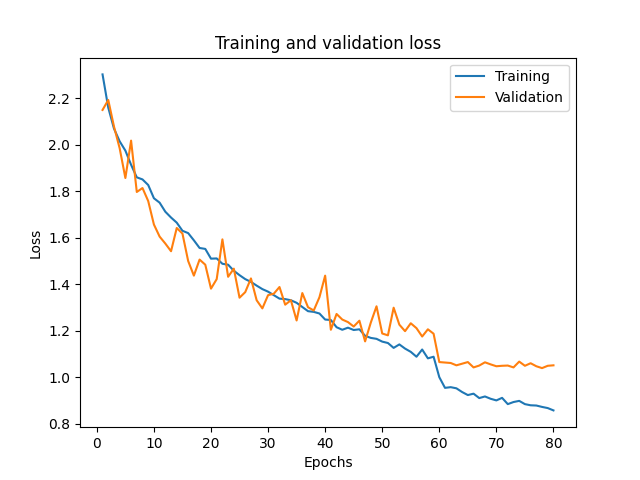}
    \caption{SGD, batch size 32}
    \label{fig:02}
  \end{subfigure}\hfill
\begin{subfigure}[b]{0.24\textwidth}
    \includegraphics[width=\textwidth]{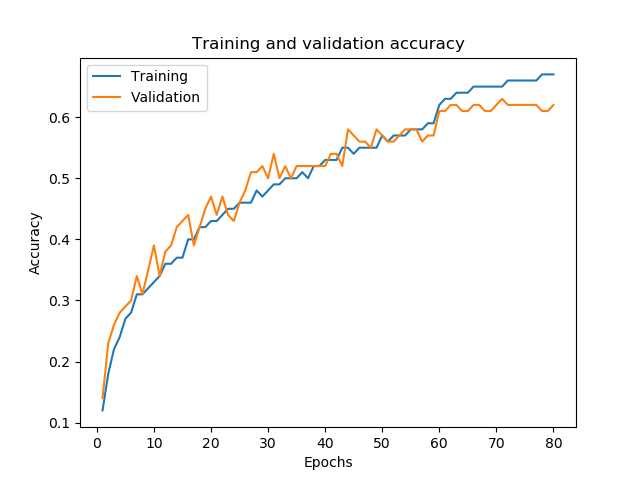}
    \caption{Adam, batch size 32}
    \label{fig:03}
  \end{subfigure}\hfill
  \begin{subfigure}[b]{0.24\textwidth}
    \includegraphics[width=\textwidth]{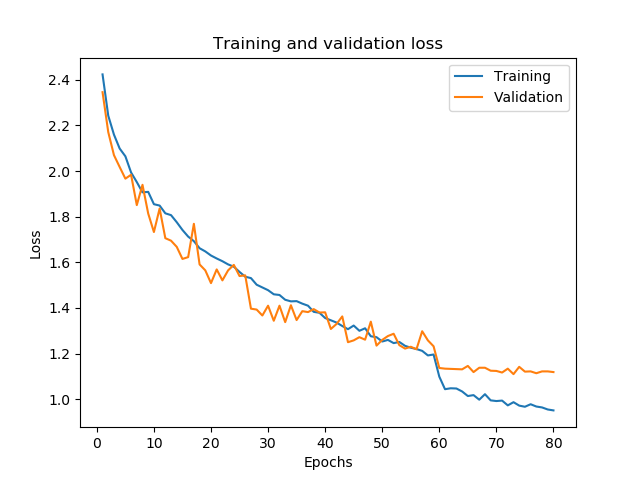}
    \caption{Adam, batch size 32}
    \label{fig:04}
  \end{subfigure}\hfill
  \begin{subfigure}[b]{0.24\textwidth}
    \includegraphics[width=\textwidth]{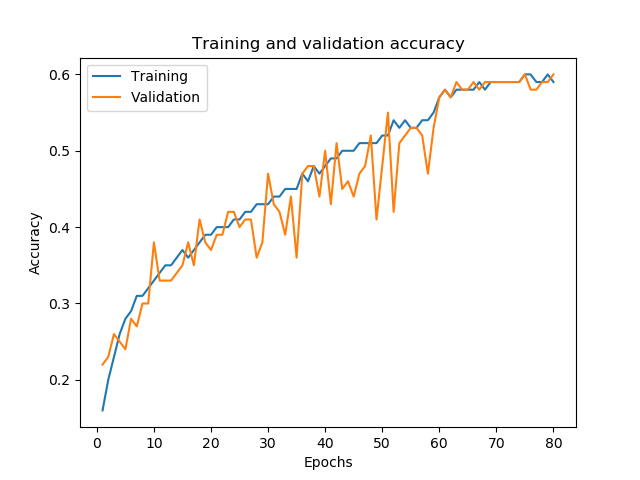}
    \caption{ASGD, batch size 32}
    \label{fig:05}
  \end{subfigure}\hfill
  \begin{subfigure}[b]{0.24\textwidth}
    \includegraphics[width=\textwidth]{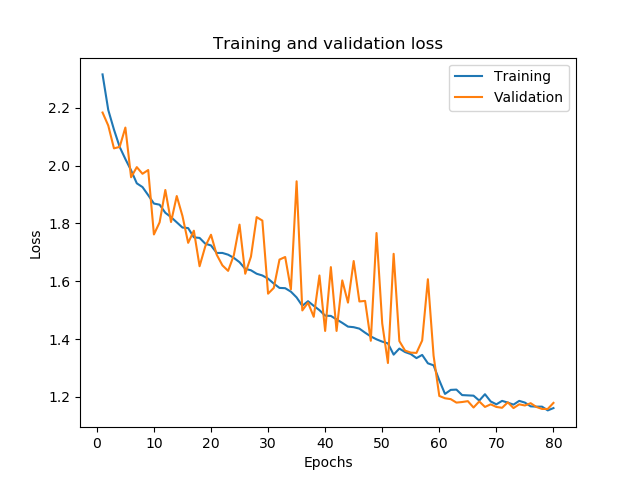}
    \caption{ASGD, batch size 32}
    \label{fig:06}
  \end{subfigure}\hfill
  \caption{Comparison of different optimizers}
  \label{fig:optimizers}

\end{figure}

We have tried SGD, Adam and ASGD optimizers to find the best one for us. Figure \ref{fig:optimizers} shows the comparison among their performances. Ideally, a model should have almost same training and validation result. In case of ASGD, the validation performance is very fluctuating. SGD has slightly higher accuracy than Adam which leads us to choose SGD optimizer.

\subsection{Learning rate}

We have tried several learning rates and found that $0.01$ works best for us. Initially, we have not used any learning rate decay. Then, we found that after $50$ epochs, the model starts overshooting. From that point, we have used $10\%$ learning rate decay after every $10$ epochs. 

\subsection{Batch size}

\begin{figure*}[ht]
  \begin{subfigure}[b]{0.25\textwidth}
    \includegraphics[width=\textwidth]{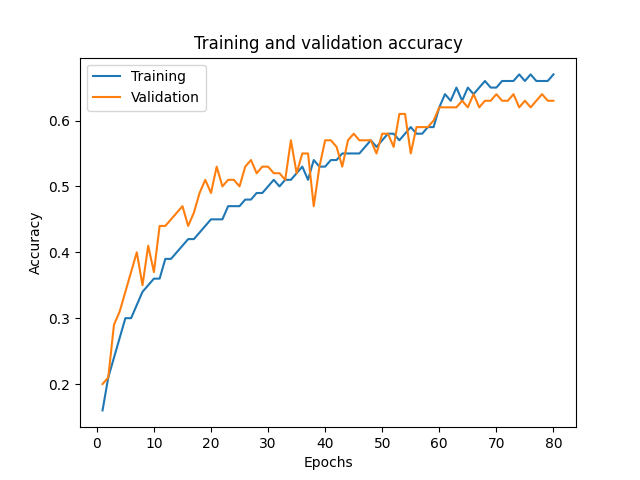}
    \caption{SGD, batch size 32}
    \label{fig:fig1}
  \end{subfigure}\hfill
  \begin{subfigure}[b]{0.25\textwidth}
    \includegraphics[width=\textwidth]{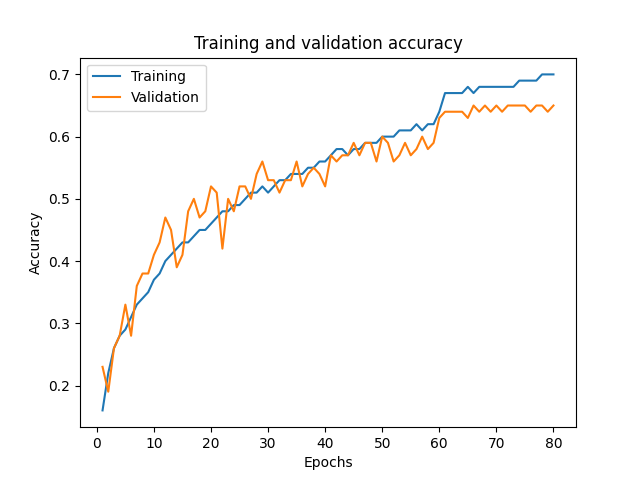}
    \caption{Adam, batch size 32}
    \label{fig:fig2}
  \end{subfigure}\hfill
  \begin{subfigure}[b]{0.25\textwidth}
    \includegraphics[width=\textwidth]{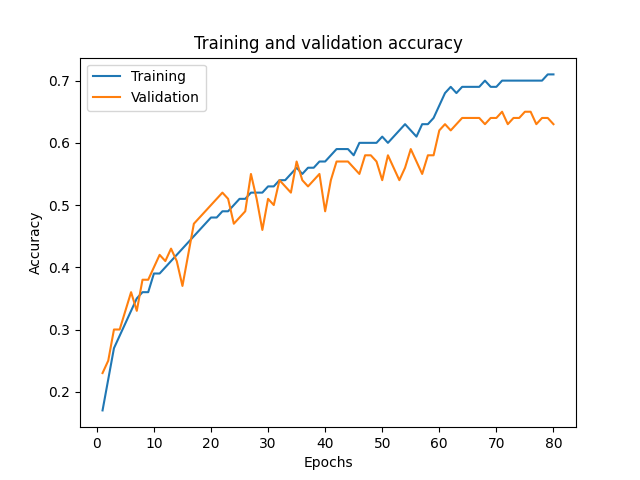}
    \caption{ASGD, batch size 32}
    \label{fig:fig3}
  \end{subfigure}\hfill
  \begin{subfigure}[b]{0.25\textwidth}
    \includegraphics[width=\textwidth]{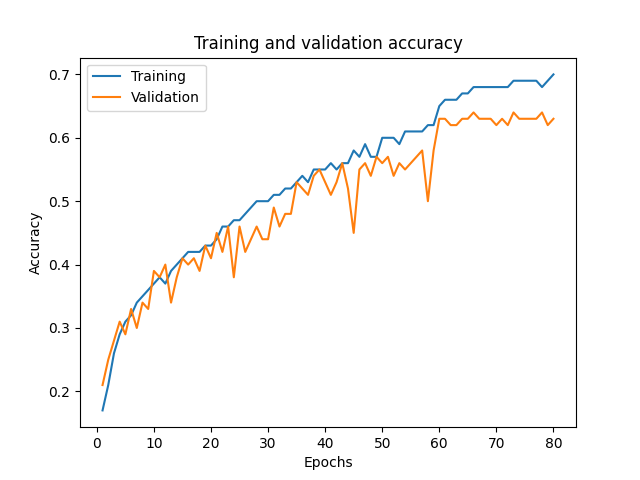}
    \caption{ASGD, batch size 32}
    \label{fig:fig4}
  \end{subfigure}\hfill
  \begin{subfigure}[b]{0.25\textwidth}
    \includegraphics[width=\textwidth]{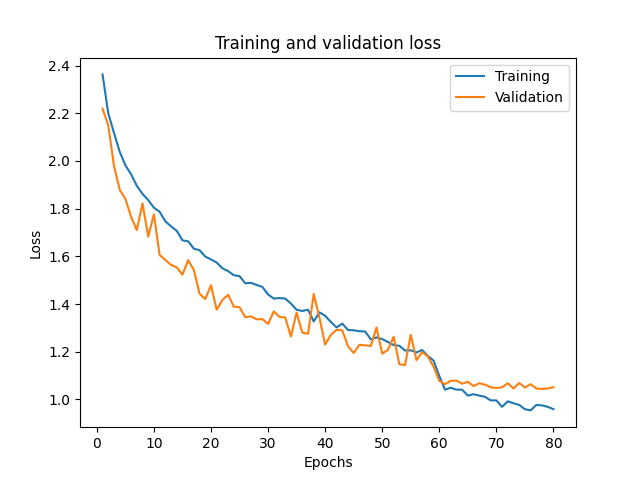}
    \caption{SGD, batch size 32}
    \label{fig:fig5}
  \end{subfigure}\hfill
  \begin{subfigure}[b]{0.25\textwidth}
    \includegraphics[width=\textwidth]{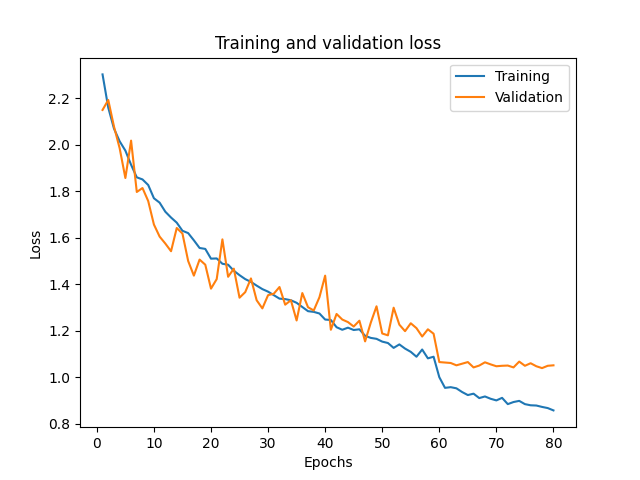}
    \caption{Adam, batch size 32}
    \label{fig:fig6}
  \end{subfigure}\hfill
  \begin{subfigure}[b]{0.25\textwidth}
    \includegraphics[width=\textwidth]{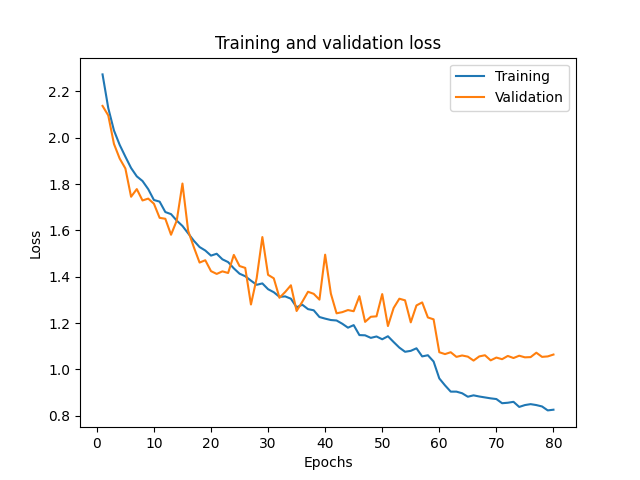}
    \caption{ASGD, batch size 32}
    \label{fig:fig7}
  \end{subfigure}\hfill
  \begin{subfigure}[b]{0.25\textwidth}
    \includegraphics[width=\textwidth]{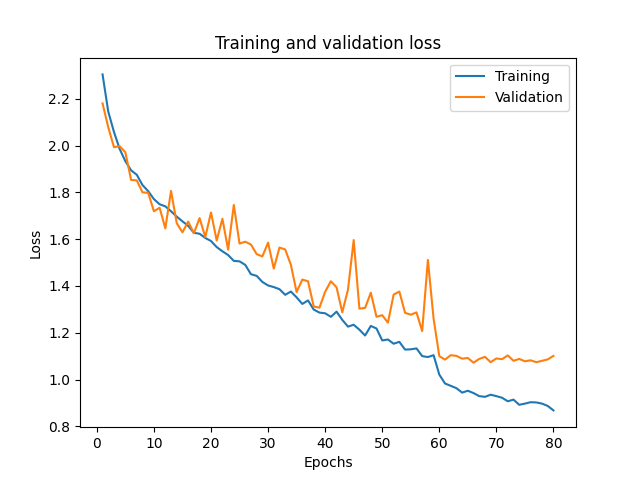}
    \caption{ASGD, batch size 32}
    \label{fig:fig8}
  \end{subfigure}\hfill
  \caption{Comparison of different batch sizes}
  \label{fig:batch}
\end{figure*}

We have tried batch size of 16, 32, 64 and 128. Figure \ref{fig:batch} presents the training and validation result for different batch size. To precisely analyse the result, we have organized the result in table \ref{table:batch}. It is noticeable that batch size does not have any significant impact on the validation accuracy or loss. We have selected batch size 32 only based on its slightly lower validation loss. 

\begin{table}[ht]
\centering
\captionof{table}{Training and validation result for different batch sizes}
\begin{tabular}{|c|c|c|c|c|}
\hline
\multirow{2}{*}{\textbf{Batch size}} & \multicolumn{2}{|c|}{\textbf{Training}} & \multicolumn{2}{|c|}{\textbf{Validation}}\\ \cline{2-5}
     & Accuracy & Loss & Accuracy & Loss\\ \hline
$16$ & $64\%$ & $1.022$  & $64\%$ & $1.056$  \\ \hline
$32$ & $67\%$ & $0.929$  & $65\%$ & $1.042$  \\ \hline
$64$ & $70\%$ & $0.854$  & $65\%$ & $1.044$ \\ \hline
$128$ & $68\%$ & $0.942$  & $64\%$ & $1.072$ \\ \hline
\end{tabular}
\label{table:batch}
\end{table}

\subsection{Dropout factor}
Dropout factor is a very important hyperparameter while designing a CNN. Higher dropout factor may lead to lower accuracy and lower dropout factor may lead to overfitting. We have carefully chosen our dropout factor $0.5$ by analyzing the performance of different dropout factors. Table \ref{table:dropout} shows the comparative performance based on dropout factor. The model has higher training accuracy but lower validation accuracy when the dropout factor is $0.1$. On the other hand, it has lower accuracy for both training and validation when the dropout factor is $0.9$.

\begin{table}[ht]
\centering
\captionof{table}{Training and validation result for different dropout factors}
\begin{tabular}{|c|c|c|c|c|}
\hline
\multirow{2}{*}{\textbf{Dropout factor}} & \multicolumn{2}{|c|}{\textbf{Training}} & \multicolumn{2}{|c|}{\textbf{Validation}}\\ \cline{2-5}
     & Accuracy & Loss & Accuracy & Loss\\ \hline
$0.1$ & $89\%$ & $0.343$  & $58\%$ & $1.507$  \\ \hline
$0.3$ & $79\%$ & $0.604$  & $65\%$ & $1.136$  \\ \hline
$0.5$ & $67\%$ & $0.929$  & $65\%$ & $1.042$  \\ \hline
$0.7$ & $57\%$ & $1.237$  & $61\%$ & $1.132$ \\ \hline
$0.9$ & $29\%$ & $1.967$  & $38\%$ & $1.778$ \\ \hline
\end{tabular}
\label{table:dropout}
\end{table}

\subsection{Snapshot Ensemble}

This is a technique that can improve the accuracy by a few percentage. There are a few ways to adapt the idea. In our case, we have summed the output of softmax layers from multiple models. Let, $P=[x_1, x_2, ..., x_11]$ and $Q=[y_1, y_2, ..., y_11]$ are the output from two models. We have observed that when a model is not confident about its prediction, it produces similar value in multiple units in the softmax layer. Our idea was to overcome this dilemma by computing $P+Q = [x_1 + y_1, x_2 + y_2, x_3 + y_3]$. Then, we predicted the output by taking the index with the maximum value in $P+Q$. It improved our validation accuracy by $2\%$. Instead of summing up the softmax output, it is also possible to independently predict the output by each model and select the winner by majority voting.  

\subsection{Performance Analysis}
\begin{figure*}[ht]
  \begin{subfigure}[b]{0.48\textwidth}
    \includegraphics[width=\textwidth]{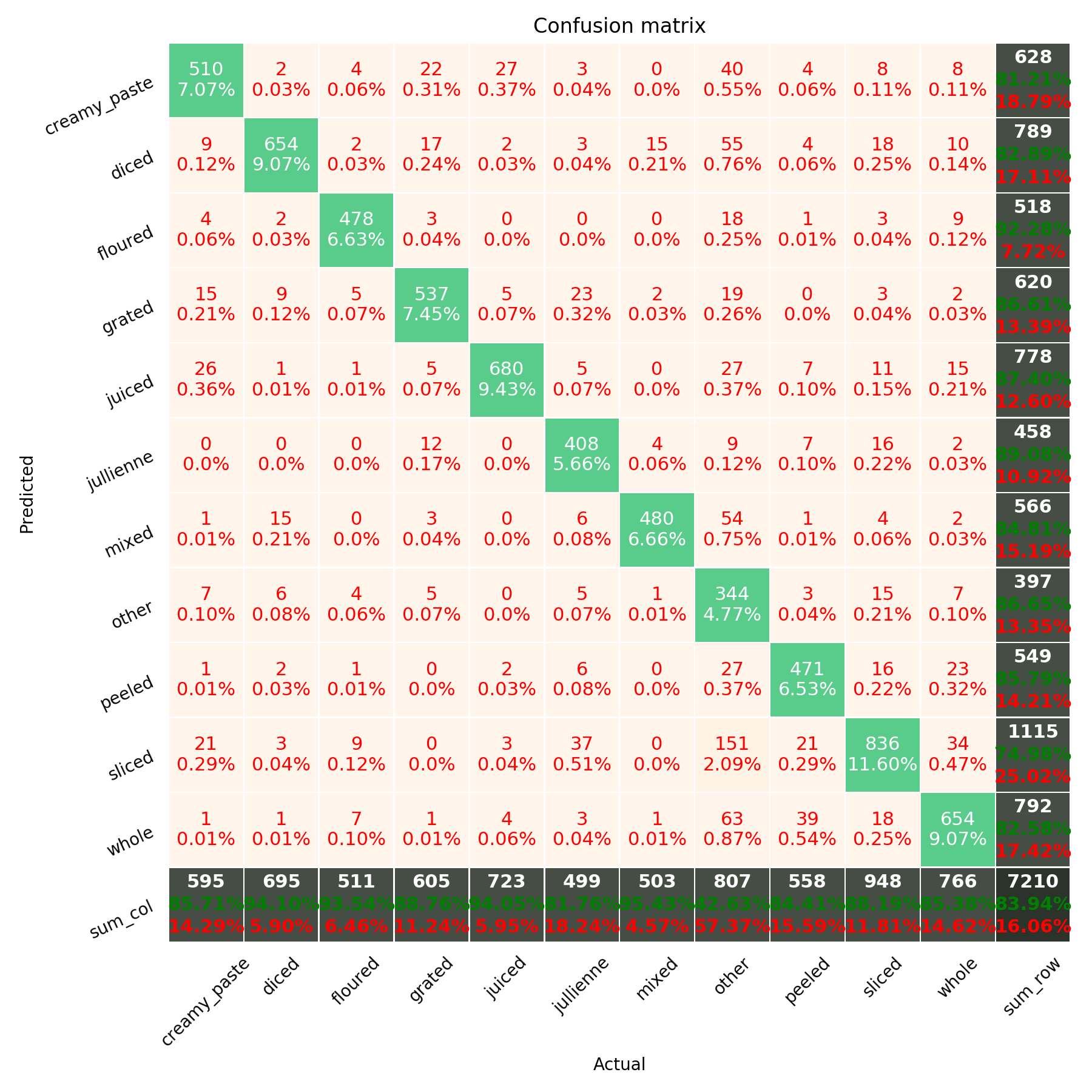}
    \caption{}
    \label{fig:11}
  \end{subfigure}\hfill
  \begin{subfigure}[b]{0.48\textwidth}
    \includegraphics[width=\textwidth]{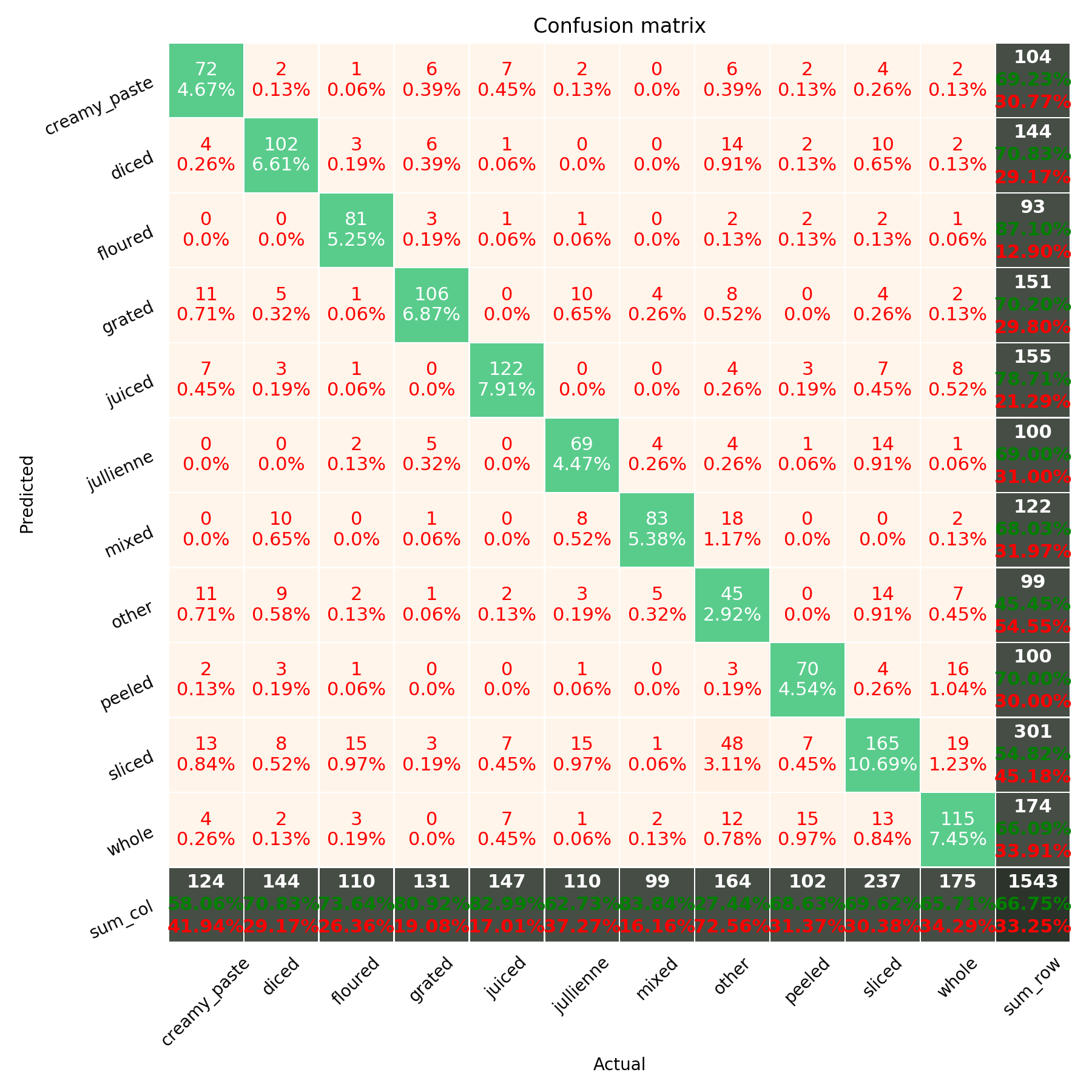}
    \caption{}
    \label{fig:22}
  \end{subfigure}
  \caption{Confusion matrix on (a) training and (b) validation dataset}
  \label{fig:confusion_matrix}
\end{figure*}

Our model achieves \textbf{66.75\%} on 1543 validation images. The confusion matrices figure \ref{fig:confusion_matrix} indicates that the model has highest accuracy for ``mixed" state and lowest accuracy for ``other" state for both training and validation dataset. We also observe that, there are 948 sliced states in the training dataset, but the model predicted total 1115 states as sliced. Same phenomenon is also visible in validation set. Since, the ``sliced" state appear more frequently in the dataset, the model created a bias towards that state. Our CNN model achieves \textbf{65.8\%} accuracy in the unseen training dataset of cooking state recognition challenge. 

%% file: discussion.tex
\section{Discussion}

\begin{figure*}[ht]
  \begin{subfigure}[b]{0.24\textwidth}
    \includegraphics[height=4cm,width=\textwidth]{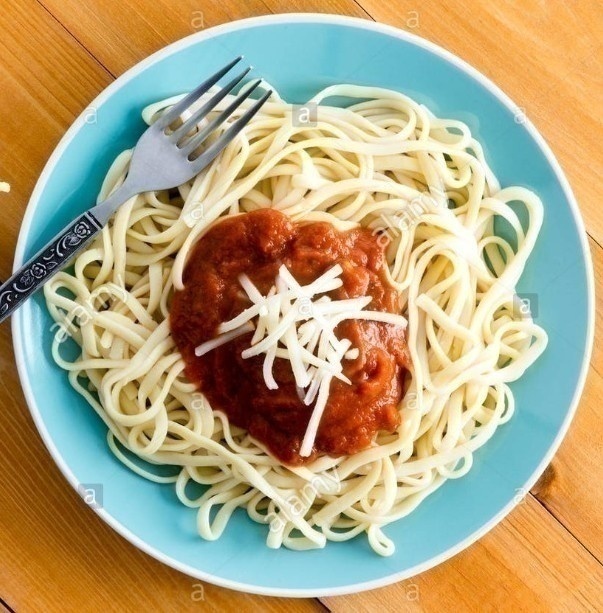}
    \caption{Predicted as ``jullienne"}
    \label{fig:0}
  \end{subfigure}
  \begin{subfigure}[b]{0.24\textwidth}
    \includegraphics[height=4cm,width=\textwidth]{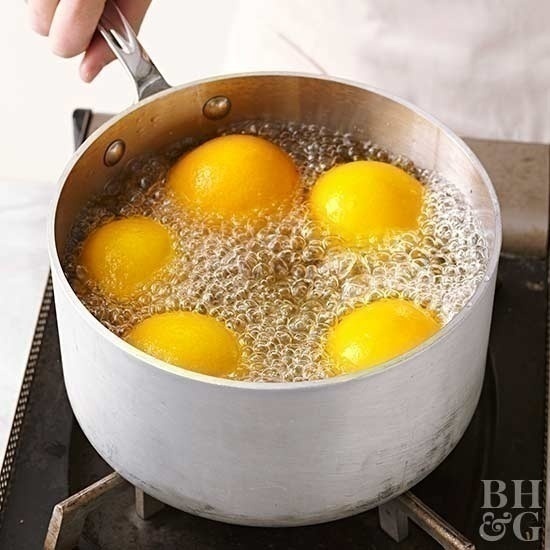}
    \caption{Predicted as ``whole"}
    \label{fig:1}
  \end{subfigure}
  \begin{subfigure}[b]{0.24\textwidth}
    \includegraphics[height=4cm,width=\textwidth]{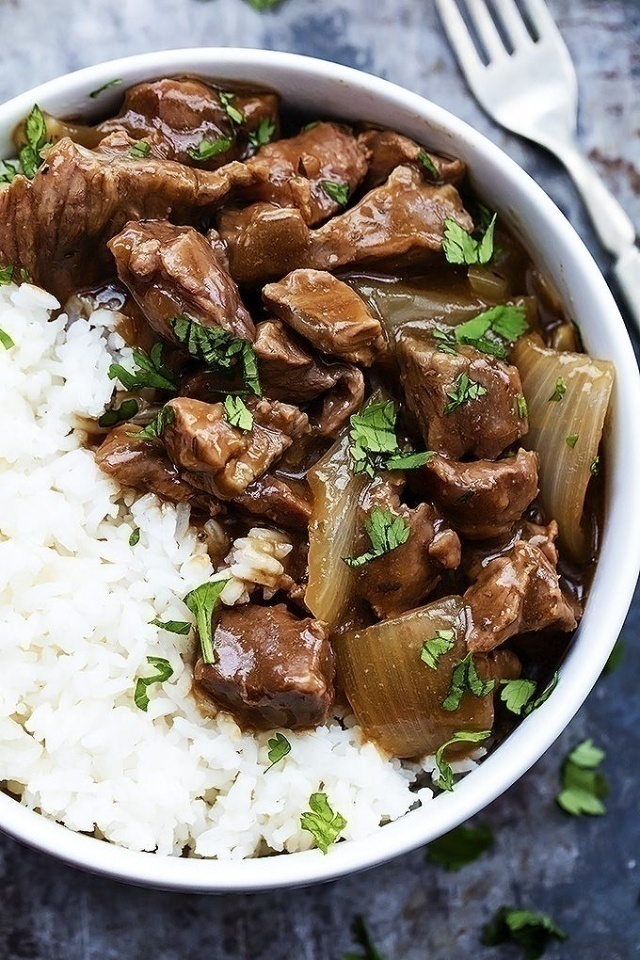}
    \caption{Predicted as ``mixed"}
    \label{fig:2}
  \end{subfigure}
  \begin{subfigure}[b]{0.24\textwidth}
    \includegraphics[height=4cm,width=\textwidth]{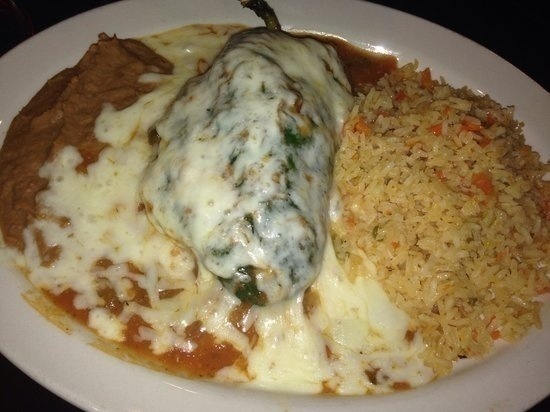}
    \caption{Predicted as ``creamy\_paste"}
    \label{fig:3}
  \end{subfigure}

  \begin{subfigure}[b]{0.24\textwidth}
    \includegraphics[height=4cm,width=\textwidth]{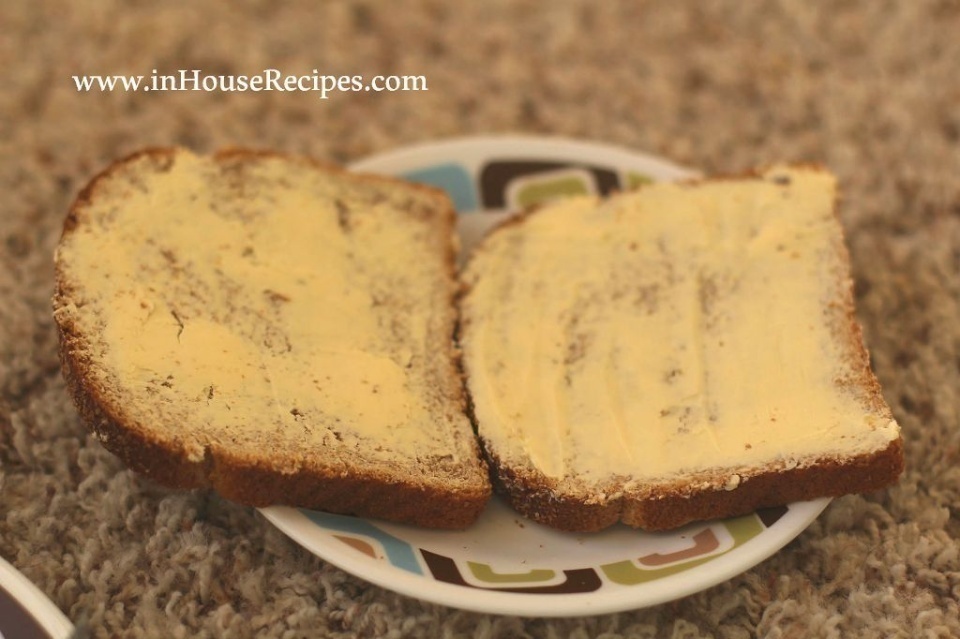}
    \caption{Predicted as ``Sliced"}
    \label{fig:4}
  \end{subfigure}
  \begin{subfigure}[b]{0.24\textwidth}
    \includegraphics[height=4cm,width=\textwidth]{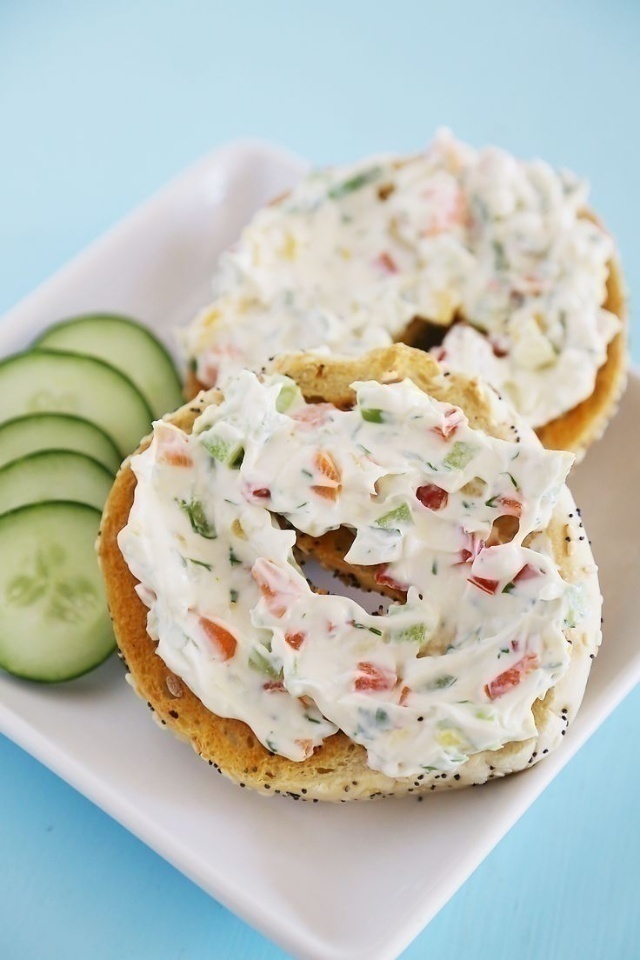}
    \caption{Predicted as ``creamy\_paste"}
    \label{fig:5}
  \end{subfigure}
  \begin{subfigure}[b]{0.24\textwidth}
    \includegraphics[height=4cm,width=\textwidth]{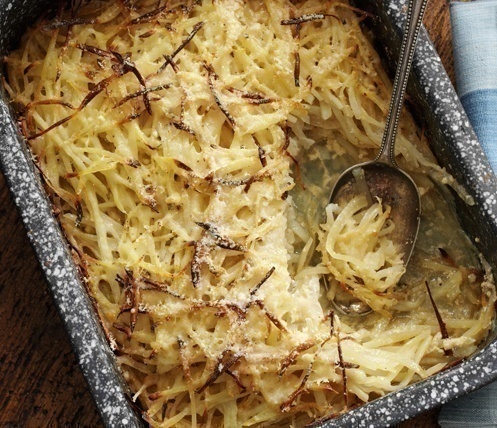}
    \caption{Predicted as ``jullienne"}
    \label{fig:6}
  \end{subfigure}
  \begin{subfigure}[b]{0.24\textwidth}
    \includegraphics[height=4cm,width=\textwidth]{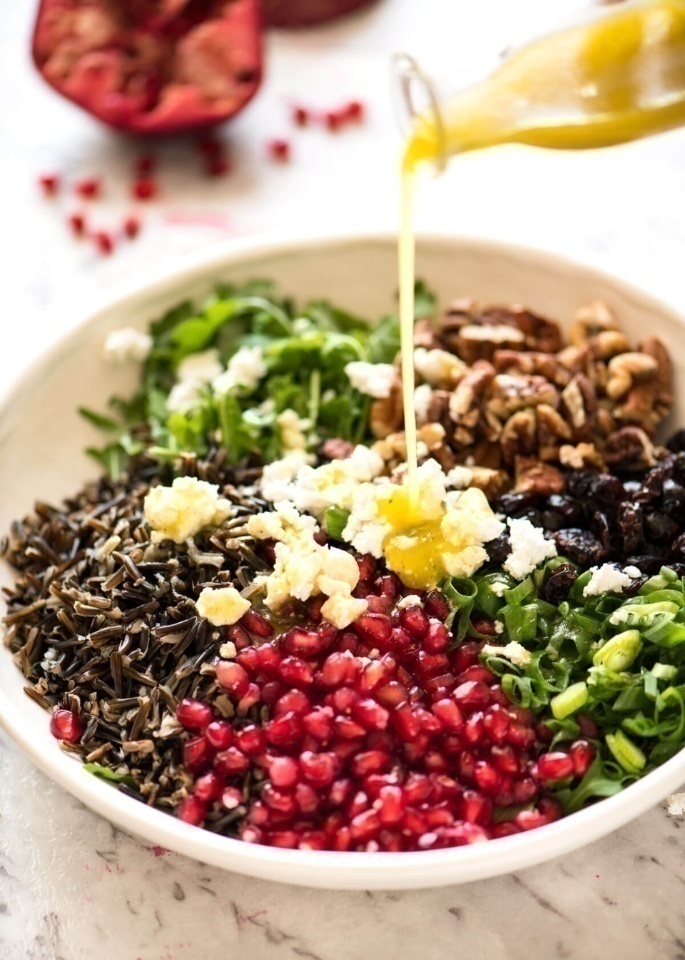}
    \caption{Predicted as ``mixed"}
    \label{fig:7}
  \end{subfigure}
  \caption{Misclassified examples, labelled as ``other" in the (a-d) training and (e-h) validation dataset}
  \label{fig:misclassified-valid}
\end{figure*}

In summary, in this paper, we have developed a CNN without using any pretrained model to classify object's state related to cooking. We have analyzed the performance based on different hypermeters and confusion matrix. It caught our attention that the model heavily struggles to correctly predict the ``other" state. It only predicted 27.44\% correctly of all the ``other" states in the validation dataset. Upon investigation, we found that, some of the images are ambiguous in terms of state. Those could be easily labelled with the state predicted by the model. Some of the misclassified examples are presented in \ref{fig:misclassified-valid}.

As future work, we will design a network adapting ideas from ResNet and Inception architecture so that the model can itself decide on the number of layers and filter sizes. We will also test the performance on other dataset. The performance should be improved if the model is trained on a larger dataset.